\documentclass{ifacconf}

\usepackage{graphicx}      % include this line if your document contains figures
\usepackage{natbib}        % required for bibliography

\usepackage{amsmath}
\usepackage{amssymb}
\usepackage{xcolor}
\usepackage{algorithm}
\usepackage{algpseudocode}
\usepackage{natbib}
\usepackage[ruled,vlined,algo2e]{algorithm2e}

\SetKwInput{KwInput}{Input}                % Set the Input
\DeclareMathOperator*{\minimize}{minimize}

\begin{document}
\begin{frontmatter}

\title{Time-optimal Point-to-point Motion Planning: A Two-stage Approach\thanksref{footnoteinfo}} 
% Title, preferably not more than 10 words.

\thanks[footnoteinfo]{This work has been carried out within the European Union’s Horizon 2020 research and innovation programme under the Marie Skłodowska-Curie grant agreement ELO-X No.953348.}

\author[First]{Shuhao Zhang} 
\author[First]{Jan Swevers} 

\address[First]{
% MECO Research Team, Department of Mechanical Engineering,\\ and Flanders Make@KU Leuven, Core lab MPRO, Leuven, Belgium.\\(e-mail: firstname.lastname@kuleuven.be)
MECO Research Team, Department of Mechanical Engineering,\\
KU Leuven, Belgium (e-mail: firstname.lastname@kuleuven.be)\\
and Flanders Make@KU Leuven, Core lab MPRO,
Leuven, Belgium
}

\begin{abstract}                % Abstract of 50--100 words
This paper proposes a two-stage approach to formulate the time-optimal point-to-point motion planning problem, involving a first stage with a fixed time grid and a second stage with a variable time grid.
The proposed approach brings benefits through its straightforward optimal control problem formulation with a fixed and low number of control steps for manageable computational complexity and the avoidance of interpolation errors associated with time scaling, especially when aiming to reach a distant goal.
Additionally, an asynchronous nonlinear model predictive control (NMPC) update scheme is integrated with this two-stage approach to address delayed and fluctuating computation times, facilitating online replanning. 
The effectiveness of the proposed two-stage approach and NMPC implementation is demonstrated through numerical examples centered on autonomous navigation with collision avoidance.

\end{abstract}


\end{frontmatter}
%===============================================================================
\section{Introduction}
Time-optimal point-to-point motion planning, which involves transitioning the system from its current state to a desired terminal state in the shortest time while continuously satisfying stage constraints, finds wide applications in various fields such as robot manipulation, cranes, and autonomous navigation.
This problem can be approached in a two-level manner, comprising high-level geometric path planning and lower-level path following considering system dynamics (\cite{Verscheure09}). 
Alternatively, it can be directly solved in the system's state space, known as the direct approach (\cite{8264024}).
The direct approach commonly formulates the time-optimal motion planning problem as a receding horizon Optimal Control Problem (OCP) and is solved either offline or repeatedly in a Nonlinear Model Predictive Control (NMPC) implementation (\cite{Zhao04}).

In this paper, our focus is on direct approaches. 
One approach proposed in \cite{7331052} involves scaling the continuous-time system with a temporal factor before discretizing it.
The OCP formulated in this time scaling approach has a variable time grid with a fixed horizon length.
The temporal factor, treated as an additional decision variable, is minimized in the objective to achieve time-optimality.
% A work implementing this time scaling approach in the context of truck-trailer autonomous mobile robot parking maneuver planning is demonstrated in \cite{BOS20234877}.
When applied with a relatively small horizon length, the computational complexity remains manageable.
Yet, the time grid is typically coarse when the time needed to reach the terminal state is long, leading to a correspondingly coarse motion trajectory.
In accordance with the discrete-time control system, the resulting motion trajectory must undergo interpolation to align appropriately. 
Regrettably, this refinement process may give rise to infeasibility concerns attributed to interpolation errors.
An alternative approach proposed by \cite{8264024} chooses to use a fixed time grid, i.e., discretizing the system with the control sampling time.
This approach, referred to as the exponential weighting approach, opts for a fixed but much larger horizon length and minimizes the L1-norm of the deviation from the desired terminal state, weighted by exponentially increasing weights, to achieve time-optimality.
A limitation of the exponential weighting approach arises when the system transitions to a distant terminal state. 
The considerable horizon length poses computational challenges for real-time implementation and introduces numerical ill-conditions due to the exponentially increasing weights.

We propose a two-stage approach to formulate the time-optimal OCP. 
This approach involves a first stage with a fixed time grid corresponding to the control sampling time and a second stage with a variable time grid derived from the time-scaled system.
It leverages the advantages of both aforementioned approaches by formulating the time-optimal OCP with a fixed and low number of control steps for computational manageability and preempting interpolation errors in the first stage.
Solving the OCP formulated through this two-stage approach using the classical NMPC implementation scheme — specifically, solving the OCP within a single control sampling time and applying the first optimal control from stage 1 to the system — can still be challenging, particularly in scenarios characterized by complex system models and stage constraints.
To ensure complete convergence in every NMPC iteration, we employ the ASAP-MPC update strategy, an asynchronous NMPC implementation scheme (\cite{ASAP-MPC}) that is designed to handle the fluctuating computational delays.

\textit{Paper structure:} In Section \ref{Sec_II}, we introduce the time-optimal point-to-point motion planning problem and discuss two approaches to formulate it: the time scaling approach and the exponential weighing approach.
In Section \ref{Sec_III}, we state the proposed two-stage approach and subsequently present a scheme that integrates this two-stage approach with the ASAP-MPC update strategy to handle fluctuating computation delays.
In Section \ref{Sec_IV}, we compare the three approaches and showcase the presented NMPC scheme through numerical examples of autonomous navigation while avoiding collisions with obstacles.
Section \ref{Sec_V} concludes the paper.

\textit{Notation:} The set of positive integer numbers is denoted by $\mathbb{N}_+$.
The L1-norm of a variable $s$ is denoted by $\|s\|_1$.
The sequence of a variable $s$ is denoted by 
$\{s\}_{n=0}^N:=s_0,s_1,...,s_N$. 

\section{Problem Setup and Preliminaries}\label{Sec_II}
We consider the following continuous-time nonlinear system:
\begin{equation}
    \frac{\text{d}s(t)}{\text{d}t}
    =f_c(s(t), u(t)),\quad t\in[0,T],
    \label{eq: continuous_system}
\end{equation}
where $t$ is the time, $s(t)\in\mathbb{R}^{n_s}$ and $u(t)\in\mathbb{R}^{n_u}$ are state and control input, respectively. 

The continuous-time time-optimal motion planning problem, which plans a feasible trajectory to transition the system (\ref{eq: continuous_system}) from an initial state $s_{\text{t0}}:=s(t_0)$ to a desired terminal state $s_{\text{tf}}:=s(t_f)$ under some stage constraints in the shortest time, is formulated as follows:
\begin{equation}
    \begin{alignedat}{2}
        \minimize_{\substack{T, s(\cdot), u(\cdot)}} &\:\int_0^T1\text{d}t &\:\\
        \text{subject to }\:s(0)&=s_{\text{t0}}&\:\\
        \:\dot{s}(t)&=f_c(s(t), u(t)),&\:\text{for } t\in[0,T]\\
        \:h(s(t),u(t))&\leq0,&\:\text{for } t\in[0,T]\\
        \:s(T)&=s_{\text{tf}}&\:\\
        \:0&\leq T,
        \label{ocp: continuous_time}
    \end{alignedat}
\end{equation}
in which $h: \mathbb{R}^{n_s}\times\mathbb{R}^{n_u}\to\mathbb{R}^{n_h}$ denotes the stage constraints. 
In the context of problem (\ref{ocp: continuous_time}), the decision variable $T$ for the system (\ref{eq: continuous_system}) represents the total
time required to transition from the initial state $s_{\text{t0}}$ to the desired terminal state $s_{\text{tf}}$.

In this paper, we are interested in solving a discretized version of the problem (\ref{ocp: continuous_time}) online with a receding planning horizon in a NMPC implementation. 
Therefore, two different approaches: the time scaling approach (\cite{7331052}), and the exponential weighting approach (\cite{8264024}), are discussed in Section \ref{Sec_II1}, and Section \ref{Sec_II2}, respectively.
\subsection{Time Scaling Approach}\label{Sec_II1}
We use a multiple shooting method with a fixed number $N$ of shooting intervals to discretize the continuous-time system (\ref{ocp: continuous_time}). 
Since the total time $T$ is not known a priori, we introduce a time scaler $\tau:=Nt/T$ such that the system (\ref{ocp: continuous_time}) becomes
\begin{equation}
    \frac{\text{d}s(t)}{\text{d}\tau}
    =f_c(s(t), u(t))\frac{T}{N},\quad \tau\in[0,N].
    \label{eq: continuous_system_scaled}
\end{equation}
This technique — time scaling — makes the total time $T$ independent of the time scaler $\tau$ over which we integrate the continuous-time system.
The problem (\ref{ocp: continuous_time}) discretized by the time scaling approach is formulated as follows:
\begin{equation}
    \begin{alignedat}{2}
        \minimize_{\substack{T, \{s\}_{n=0}^N, \{u\}_{n=0}^{N-1}}} &\: T &\:\\
        \text{subject to }\:s_0&=s_{\text{t0}}&\:\\
        \:s_{n+1}&=f_{T}(s_n, u_n,\Delta T),&\:\text{for } n\in[0,N-1]\\
        \:h(s_n,u_n)&\leq0,&\:\text{for } n\in[0,N-1]\\
        \:s_N&=s_{\text{tf}}&\:\\
        \:0&\leq T,
        \label{ocp: time_scaling}
    \end{alignedat}
\end{equation}
where $\Delta T:=T/N$ denotes the temporal discretization, and the function $s_{n+1}=f_{T}(s_n, u_n,\Delta T)$ denotes the discrete-time representation of the time-scaled system (\ref{eq: continuous_system_scaled}), which is obtained by numerical integration. 
\subsection{Exponential Weighting Approach}\label{Sec_II2}
Unlike the time scaling approach, which involves a fixed horizon length $N$ and optimizes the total time $T$, the second approach employs a discrete-time model 
\begin{equation}
    s_{n+1}
    =f_d(s_{n}, u_{n}),\quad n=0,1,...,
    \label{eq: discrete_time}
\end{equation}
derived by numerical integrating the continuous-time system (\ref{eq: continuous_system}) over a fixed sampling interval $t_s$, which also serves as the control sampling time.

One time-optimal OCP using the discrete-time model (\ref{eq: discrete_time}) is defined as below:
\begin{equation}
    \begin{alignedat}{2}
        N^*(s_{\text{t0}},s_{\text{tf}}):=\quad\quad\quad&&\\
        \minimize_{\substack{N, \{s\}_{n=0}^N, \{u\}_{n=0}^{N-1}}} &\: N &\:\\
        \text{subject to }\:s_0&=s_{\text{t0}}&\:\\
        \:s_{n+1}&=f_{d}(s_n, u_n),&\:\text{for } n\in[0,N-1]\\
        \:h(s_n,u_n)&\leq0,&\:\text{for } n\in[0,N-1]\\
        \:s_N&=s_{\text{tf}}&\:\\
        \:N&\in\mathbb{N}_+,
        \label{ocp: vairable_N}
    \end{alignedat}
\end{equation}
which is a mixed-integer programming problem.
It finds the minimal $N^*(s_{\text{t0}},s_{\text{tf}})\in\mathbb{N}_+$ to transition the discrete-time model (\ref{eq: discrete_time}) from the initial state $s_{\text{t0}}$ to the desired terminal state $s_{\text{tf}}$. 

Since the horizon length $N$ in the problem (\ref{ocp: vairable_N}) is a decision variable that is not fixed throughout the NMPC implementation, this can constantly change the size of the OCP to be solved, and is therefore inconvenient in real-time execution.
A more convenient reformulation of the time-optimal OCP uses a fixed $N$ that is larger than $N^*(s_{\text{t0}},s_{\text{tf}})$, and minimizes the weighted L1-norm of the difference between the state at each shooting point $s_n$ and the desired terminal state $s_{\text{tf}}$ with exponentially increased weighting factors.
This OCP proposed in \cite{8264024} is presented as follows:
\begin{equation}
    \begin{alignedat}{2}
        \minimize_{\substack{\{s\}_{n=0}^N, \{u\}_{n=0}^{N-1}}} &\: \sum_{n=0}^{N-1}\gamma^n\|s_n-s_{\text{tf}}\|_1 &\:\\
        \text{subject to }\:s_0&=s_{\text{t0}}&\:\\
        \:s_{n+1}&=f_{d}(s_n, u_n),&\:\text{for } n\in[0,N-1]\\
        \:h(s_n,u_n)&\leq0,&\:\text{for } n\in[0,N-1]\\
        \:s_N&=s_{\text{tf}},
        \label{ocp: exponential_weight}
    \end{alignedat}
\end{equation}
where $\gamma\in\mathbb{R}>1$ is a fixed pre-defined parameter. 
Note that choosing the L1-norm of the state difference induces sparsity, that is yielding that some components of the objective are exactly zero. 
Consequently, at a later stage than $N^*(s_{\text{t0}},s_{\text{tf}})$, the equality $s_{n+1}=s_n$ holds so that the state $s_n$ is stabilized to the desired terminal state $s_{\text{tf}}$, and the solution will be time-optimal.
\subsection{Discussion}
Both approaches to formulate the time-optimal motion planning problem are approximations of the continuous-time problem (\ref{ocp: continuous_time}).
These approximations are made under the condition that the control inputs are piecewise constant parameterized, and the time grid remains evenly spaced over the total horizon with a fixed number of control steps.
The time scaling approach is more directly linked to the problem (\ref{ocp: continuous_time}) as it minimizes the total time $T$. 
Even in scenarios where reaching a distant desired terminal state $s_{\text{tf}}$ requires a considerable amount of time $T$, the computational overhead remains manageable.
However, large total time $T$ with a fixed horizon length $N$ may result in a coarse time grid, introducing a safety concern as the stage constraints are only activated at the shooting points, and the time-optimal solution often lies at the edge of these constraints.
In a NMPC implementation, for example, the problem (\ref{ocp: time_scaling}) needs to be solved repeatedly. 
\cite{BOS20234877} interpolates the time-optimal solutions of the problem (\ref{ocp: time_scaling}) with the control sampling time $t_s$ to apply the optimal control to the system. 
Yet, the interpolation introduces errors that may lead to infeasibility, e.g., the updated $s_{\text{t0}}$ for replanning is inside the obstacle.

In contrast, the exponential weighting approach derives a better and safer approximation by using a fixed but small sampling interval $t_s$.
No interpolation is needed in a NMPC implementation when applying the optimal control to the system.
Yet, when aiming to reach a distant desired terminal state $s_{\text{tf}}$, a significantly larger horizon length $N$ needs to be selected, leading to excessively increased computational complexity and numerical ill-conditions due to exponentially increased weights.
\section{Two-stage Time-optimal OCP}\label{Sec_III}
We propose a two-stage approach to formulate the time-optimal OCP, which combines the advantages of the time scaling approach and the exponential weighting approach, as follows:
\begin{equation}
    \begin{alignedat}{3}
        \minimize_{\substack{\{s_1\}_{n=0}^{N_1}, \{u_1\}_{n=0}^{N_1-1},\\\{s_2\}_{n=0}^{N_2}, \{u_2\}_{n=0}^{N_2-1},\\T_2}} &\: w_1\sum_{n=0}^{N_1-1}\gamma^n\|s_{1,n}-s_{\text{tf}}\|_1&+w_2T_2 \\
        \text{subject to }\:s_{1,0}&=s_{\text{t0}}\\
        \:s_{1,n+1}&=f_{d}(s_{1,n}, u_{1,n}),&\:\text{for } n\in&[0,N_1-1]\\
        \:h(s_{1,n},u_{1,n})&\leq0,&\:\text{for } n\in&[0,N_1-1]\\
        \:s_{1,N_1}&=s_{2,0}\\
        \:s_{2,n+1}&=f_{T}(s_{2,n}, u_{2,n},\frac{T_2}{N_2}),&\:\text{for } n\in&[0,N_2-1]\\
        \:h(s_{2,n},u_{2,n})&\leq0,&\:\text{for } n\in&[0,N_2-1]\\
        \:s_{2,N_2}&=s_{\text{tf}},
        \label{ocp: two_stage}
    \end{alignedat}
\end{equation}
where $N_1$ and $N_2$ are fixed horizon lengths of the two stages. 
$\{s_1\}_{n=0}^{N_1}$ and $\{u_1\}_{n=0}^{N_1-1}$ denote the state and control input sequences of stage 1, respectively, in which the discrete-time model (\ref{eq: discrete_time}) with the fixed sampling interval $t_s$ is used.
$\{s_2\}_{n=0}^{N_2}$ and $\{u_2\}_{n=0}^{N_2-1}$ denote the state and control input sequences of stage 2, respectively, in which the discrete-time representation $s_{2,n+1}=f_{T}(s_{2,n}, u_{2,n}, T_2/N_2)$ of the time-scaled system (\ref{eq: continuous_system_scaled}) is used and $T_2$ denotes the total time of stage 2.
The problem (\ref{ocp: two_stage}) constrains the first state of the stage 1 $s_{1,0}$ to the initial state $s_{\text{t0}}$, and the last state of the stage 2 $s_{2, N_2}$ to the terminal state $s_{\text{tf}}$. 
The last state of stage 1 is stitched to the first state of stage 2 by the equality constraint $s_{1, N_1} = s_{2,0}$.

The objective function in the problem (\ref{ocp: two_stage}) comprises two components, stemming from stages 1 and 2, respectively. 
The weighting factors, $w_1$ and $w_2$, are used to signify the relative importance of the objectives associated with the respective stages.
Note that the total time required to transition from the initial state $s_{\text{t0}}$ to the desired terminal state $s_{\text{tf}}$ in problem (\ref{ocp: two_stage}) is given by $T=N_1t_s + T_2$ when $T_2$ is positive, in other words, the total time for stage 1 remains fixed.
The optimal value of the total time $T_2$ for stage 2 approaches and becomes zero when the system is in close proximity to the desired terminal state.
Therefore, two phases should be considered when choosing the weighting factors.
In the first or initial phase, when $T_2$ remains positive, we choose $w_1$, $w_2$ such that $w_2T_2$ is much larger than $w_1\sum_{n=0}^{N_1-1}\gamma^n\|s_{1,n}-s_{\text{tf}}\|_1$ to prioritize stage 2 dominance in time optimality.
In the second or end phase, as $T_2$ approaches and becomes zero, $w_1$ and $w_2$ are updated to align the problem (\ref{ocp: two_stage}) with the problem (\ref{ocp: exponential_weight}), emphasizing stage 1 with its objective $w_1\sum_{n=0}^{N_1-1}\gamma^n\|s_{1,n}-s_{\text{tf}}\|_1$ dominance in time optimality.
We will discuss this in detail in the next subsection in the context of the NMPC implementation.
\subsection{NMPC Update Implementation}
In a classical implementation of NMPC, the OCP is solved in between every two control steps.
The first optimal control solution is applied to the system at the time instance $t_0$. 
It will then proceed to the next time instance, i.e., $t_0+t_s$, and solve the OCP again, incorporating an updated initial state and potential updated stage constraints.
This repetitive cycle continues until the system reaches the desired terminal state $s_{\text{tf}}$.
In this scenario, it is reasonable to set $N_1$ equal to 1.
However, in cases where the complex nonlinear system encounters complex stage constraints, such as collision avoidance constraints, 
the NMPC solution time may exceed the control sampling time $t_s$. 
The widely adopted Real-Time Iterations (RTI) technique, introduced by \cite{RTI}, aims to ensure high update rates by implementing only the first iteration of the OCP solution. 
However, this introduces potential safety risks, as it becomes impossible to guarantee constraint satisfaction.

A modified implementation of NMPC updates — Asynchronous NMPC (ASAP-MPC) (\cite{ASAP-MPC}) — was proposed recently to deal with the computational delays, i.e., the NMPC solution time takes more than one control sampling time $t_s$, and allows for full convergence of the optimization problem.
The ASAP-MPC strategy selects a future state $s(t_n)$ at the time instance $t_n=t_0+nt_s$ from the current solution as the initial state for the next replanning with $nt_s$ the (estimated) time required to find the OCP solution for this replanning, 
and assumes that the actual state $\hat{s}(t_n)$ at the time instance $t_n$ in the future corresponds to the predicted future state $s(t_n)$. 
This last requirement requires a (low-level) stiff tracking controller to ensure that the actual state tracks the predicted future state, i.e., $\hat{s}(t_n)\approx s(t_n)$.
Then, this predicted future state serves as the best estimate of the system's state after obtaining the replanning solution, seamlessly stitching the new solution to the previous one at this point.

\begin{algorithm2e}[t]
  \KwInput{$s_{\text{t0}}$, $s_{\text{tf}}, N_1, N_2, t_s, \gamma, w_1, w_2$}
  $\{s_1\}_{n=0}^{N_1}, \{u_1\}_{n=0}^{N_1-1}, T_2 \gets$ solve the problem (\ref{ocp: two_stage})\;
  $n_{\text{update}} = N_1$\;
  $s_{\text{t0}} \gets s_{1,n_{\text{update}}}$\;
  \While{$\mathrm{abs}(s_{\mathrm{t0}} - s_{\mathrm{tf}})\geq1e-6$}{
    \If{$T_2 - n_{\mathrm{update}}t_s \leq 0$}{
      Update $w_1, w_2$\;
    }
  $\{s_1\}_{n=0}^{N_1}, \{u_1\}_{n=0}^{N_1-1}, T_2 \gets$ solve the problem (\ref{ocp: two_stage})\;
  get the computation time $t_{\mathrm{comp}}$ of solving the problem (\ref{ocp: two_stage})\;
  $n_{\text{update}} = \text{ceil}(t_{\mathrm{comp}}/t_s)$\;
  $s_{\text{t0}} \gets s_{1,n_{\text{update}}}$\;
  }
  \caption{Two-stage time-optimal motion planning solved repeatedly in the ASAP-MPC formulation}
  \label{algo1}
\end{algorithm2e}
Here, we employ the ASAP-MPC update strategy to repeatedly solve the problem (\ref{ocp: two_stage}) for transitioning the system from the initial state $s_{\text{t0}}$ to the desired terminal state $s_{\text{tf}}$.
The combination of both is summarized in the Algorithm \ref{algo1}.
In addition to $s_{\mathrm{t0}}$ and $s_{\mathrm{tf}}$, the algorithm takes fixed values for $N_1$, $N_2$, $t_s$, $\gamma$, and assigns weighting factors $w_1$ and $w_2$ as part of its input.
This initial choice of $w_1$ and $w_2$ is to signify that stage 2 dominates time-optimality in the initial phase.
It solves the problem (\ref{ocp: two_stage}) for the first time. 
Afterward, it chooses the last state of stage 1 $s_{1, N_1}$ as the new initial state $s_{\mathrm{t0}}$ for replanning. 
Assuming accurate trajectory tracking, this selection aligns with the ASAP-MPC update strategy.
More specifically, the total time $N_1t_s$ for stage 1 is designed to represent the worst-case computation time required to solve the problem (\ref{ocp: two_stage}) so that $s_{1, N_1}$ is a just-in-time selection.
In the subsequent solves, the computation time $t_{\mathrm{comp}}$ is recorded, and the update index $n_{\mathrm{update}}\in[1, N_1]$ is computed as the smallest integer equal to or greater than $t_{\mathrm{comp}}/t_s$ for updating $s_{\text{t0}}$.
The condition $T_2 - n_{\mathrm{update}}t_s \leq 0$ indicates that the total time for the next replanning will be equal to or smaller than the trajectory time of stage 1. 
Therefore, it updates the values of $w_1$ and $w_2$ to signify that stage 1 dominates time-optimality in the end phase (stage 2 may be deemed to have ceased to exist) and stabilizes the system to the desired terminal state $s_{\mathrm{tf}}$.
\section{Numerical Example and Discussion}\label{Sec_IV}
The remainder of this paper validates the proposed two-stage approach through two numerical examples.
The first one, detailed in Section \ref{Sec_IV1}, compares the two-stage approach with the alternatives discussed in Section \ref{Sec_II}.
This comparison evaluates the time-optimal trajectory, computation time, and feasibility of the collision avoidance constraint by solving a single optimal control problem.
The second example, presented in Section \ref{Sec_IV2}, demonstrates the integration of the two-stage approach with the ASAP-MPC update strategy, addressing challenges arising from delayed and fluctuating computation times.

Both numerical examples involve time-optimal point-to-point motion planning for a point-mass unicycle model while avoiding an elliptical obstacle.
We consider the following unicycle model with position $(x,y)$ and orientation $\theta$ as its states, and the forward speed $v$ and the angular velocity $w$ as its control inputs:
\begin{equation}
    s=\begin{bmatrix}
        x\\
        y\\
        \theta
    \end{bmatrix}, 
    u=\begin{bmatrix}
        v\\
        \omega
    \end{bmatrix},
    \dot{s}(t)=\begin{bmatrix}
        v(t)\cos\theta(t)\\
        v(t)\sin\theta(t)\\
        \omega(t)
    \end{bmatrix}.
\end{equation}
% consider integration time explicitly????
The continuous-time system is discretized using the explicit Runge-Kutta method of order 4, implemented with CasADi (\cite{casadi}).
For the discrete-time model (\ref{eq: discrete_time}), the control sampling time $t_s = 0.02s$.

The time-optimal motion planning problem encompasses two types of stage constraints: control input limits 
% ($v\in[0, 0.5]\ \text{m/s}$ and $\omega\in[-\pi/3, \pi/3]\ \text{rad/s}$)
($0\leq v\leq 0.5 [\text{m/s}]$ and $-\pi/3\leq\omega\leq\pi/3 [\text{rad/s}]$)
, and collision avoidance with an elliptical obstacle defined by parameters $p_e=[x_e,y_e,a_e,b_e,\theta_e]$ as follows:
\begin{equation}
    h_e: 1-{p^{\mathrm{diff}}}^\top\Omega_ep^{\mathrm{diff}}\leq0,
    \label{eqn:collision}
\end{equation}
where $p^{\mathrm{diff}}=\begin{bmatrix}x-x_e\\y-y_e\end{bmatrix}$, and $\Omega_e=R(\theta_e)^\top\text{diag}(\frac{1}{a_e^2}, \frac{1}{b_e^2})R(\theta_e)$ with $R(\theta_e)$ represents the elliptical rotation matrix.

The two-stage time-optimal OCP and the two alternatives are formulated in Python using the Rockit toolbox for rapid OCP prototyping, presented in \cite{rockit}, and solved with Ipopt by \cite{ipopt} using ma57 of \cite{hsl} as the linear solver. 
% Key steps in the implementation including integrator and Jacobian matrix calculation are handled using Casadi \cite{casadi}.
All computations are performed on a laptop with an Intel$^{\circledR}$ Core\textsuperscript{\texttrademark} i7-1185G7 processor with eight cores at 3GHz and with 31.1GB RAM.

\subsection{Comparison of Time-Optimal Approaches}\label{Sec_IV1}
To compare OCPs formulated by the three approaches (\ref{ocp: time_scaling}), (\ref{ocp: exponential_weight}) and (\ref{ocp: two_stage}), we define the following problem, aiming to transition the system from an initial state $s_{\text{t0}}=[0.70713\mathrm{m}, 1.83274\mathrm{m}, 1.38778\mathrm{rad}]^\top$, which is a position on the edge of the elliptical obstacle, to a terminal state $s_{\text{tf}}=[4\mathrm{m}, 3.5\mathrm{m}, 0\mathrm{rad}]^\top$.
The elliptical obstacle is with parameter $p_e=[2.5\mathrm{m},1\mathrm{m},2\mathrm{m},1\mathrm{m},-\pi/6\mathrm{rad}]$.
The time scaling approach chooses a horizon length of 50.
The exponential weighting approach chooses a horizon length of 400 and the weighting factor $\gamma=1.025$.
For both stages of the two-stage approach, horizon lengths $N_1=N_2=25$ are chosen.
In addition, the two-stage approach chooses the same value of $\gamma$ as the exponential weighting approach and the weighting factor $w_1=0$ and $w_2=1$.

\begin{figure}[t]
    \begin{center}
    \includegraphics[width=8.4cm]{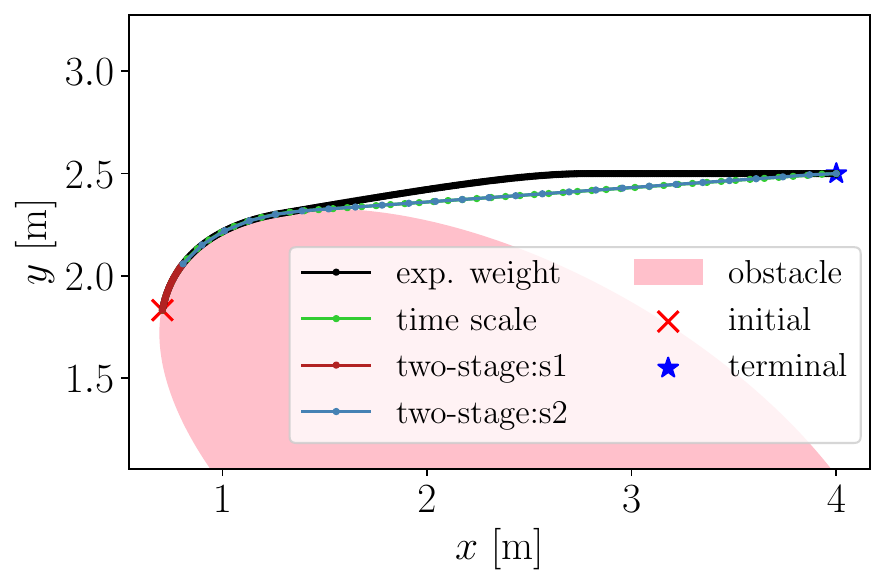}    % The printed column width is 8.4 cm.
    \caption{$x-y$ position trajectories, obtained by the three approaches.} 
    \label{fig:compare_3_traj}
    \end{center}
\end{figure}

Fig. \ref{fig:compare_3_traj} illustrates the $x-y$ position trajectory obtained by the three approaches. 
The trajectories derived from the time scaling approach and the two-stage approach are nearly identical. 
In this context, the total trajectory times for the time scaling approach and the two-stage approach are 7.0428s and 7.0439s, respectively.
The minimal horizon length for the exponential weighting approach is $N^*=353$, leading to a total trajectory time of $N^*t_s=7.06s$. The difference in the optimal trajectory times is noticeably smaller than one control sampling time $t_s$. 
The difference in motion trajectories is caused by the different optimization objectives.
Specifically, in this example, the exponential weighting approach strives to reach the terminal state sooner in the $y$ position and $\theta$ orientation than in the $x$ position, which is a direct consequence of the L1-norm objective.
This illustrates the inherent non-uniqueness of time-optimal motion planning in discrete-time.
\begin{figure}[t]
    \begin{center}
    \includegraphics[width=8.4cm]{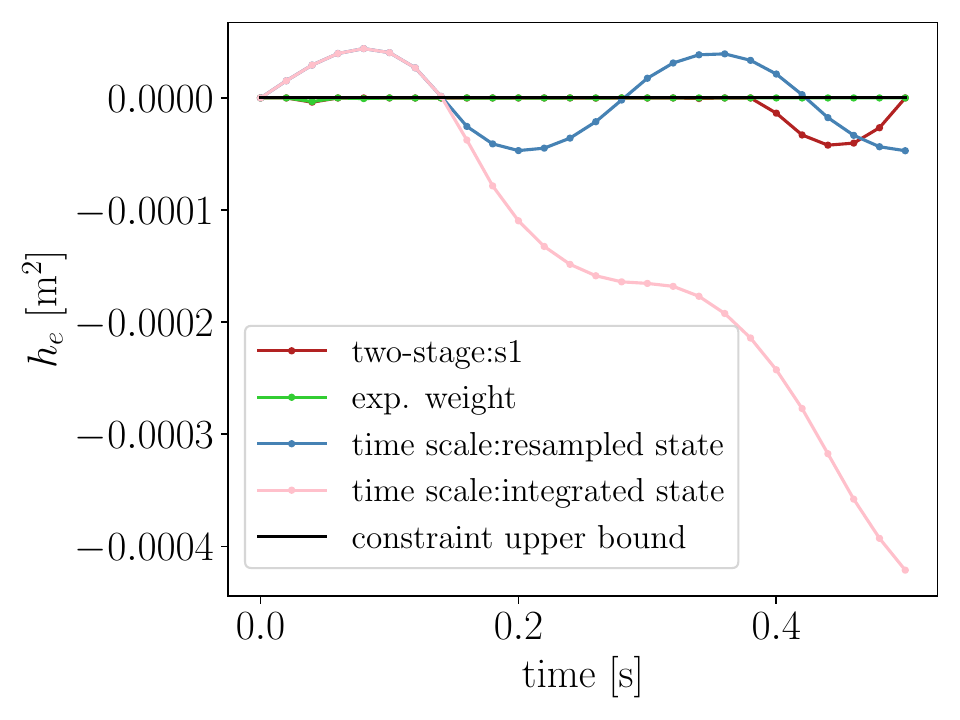}    % The printed column width is 8.4 cm.
    \caption{Compare the satisfaction of collision avoidance constraint. It is considered feasible when the value of the collision avoidance constraint does not exceed zero.} 
    \label{fig:compare_3_h}
    \end{center}
\end{figure}

In regard to computation time for this example, the time scaling approach requires approximately 0.06s, the two-stage approach takes around 0.1s, and the exponential weighting approach takes around 1.5s. 
All of these durations surpass one control sampling time, i.e., 0.02s. 
The computational overhead of the exponential weighting approach is significantly higher than that of the other two approaches, primarily due to the notably larger number of decision variables resulting from the large horizon length.

Furthermore, we showcase the satisfaction of collision avoidance constraint (\ref{eqn:collision}) during the initial 0.5 seconds (the trajectory time of stage 1 in the two-stage approach) in Fig. \ref{fig:compare_3_h}.
Both the two-stage approach and the exponential weighting approach undoubtedly meet the collision avoidance constraint.
To align with the time grid, we interpolate both state and control input trajectories obtained from the time scaling approach using the control sampling time $t_s$.
However, for the time scaling approach, both the interpolated state trajectory and the simulated state trajectory with interpolated control inputs fall short of fully satisfying the collision avoidance constraint, posing a risk of collision.

\subsection{Integration of the Two-stage Approach with the ASAP -MPC Update Strategy}\label{Sec_IV2}
This example demonstrates the integration of the two-stage approach with the ASAP-MPC update strategy, as outlined in Algorithm \ref{algo1}.
The horizon length for both stages is set to be $N_1=N_2=25$, with weighting factors $w_1 = 1$ and $w_2 = 1000$ when signifying stage 2 dominates time-optimality; otherwise, set $w_1 = 1000$ and $w_2 = 1$.
% (We choose not to set the value of the weighting factors to 0 to maintain the OCP code structure.)
The task is to transition the unicycle model from the initial state $s_{\text{t0}}=[0.1\mathrm{m}, 0.5\mathrm{m}, 0\mathrm{rad}]^\top$ to a desired terminal state $s_{\text{tf}}=[5\mathrm{m}, 2.5\mathrm{m}, 0\mathrm{rad}]^\top$, avoiding the elliptical obstacle $p_e=[2.5\mathrm{m},1\mathrm{m},2\mathrm{m},1\mathrm{m},\pi/6\mathrm{rad}]$ in the meantime.
\begin{figure}[t]
    \begin{center}
    \includegraphics[width=8.4cm]{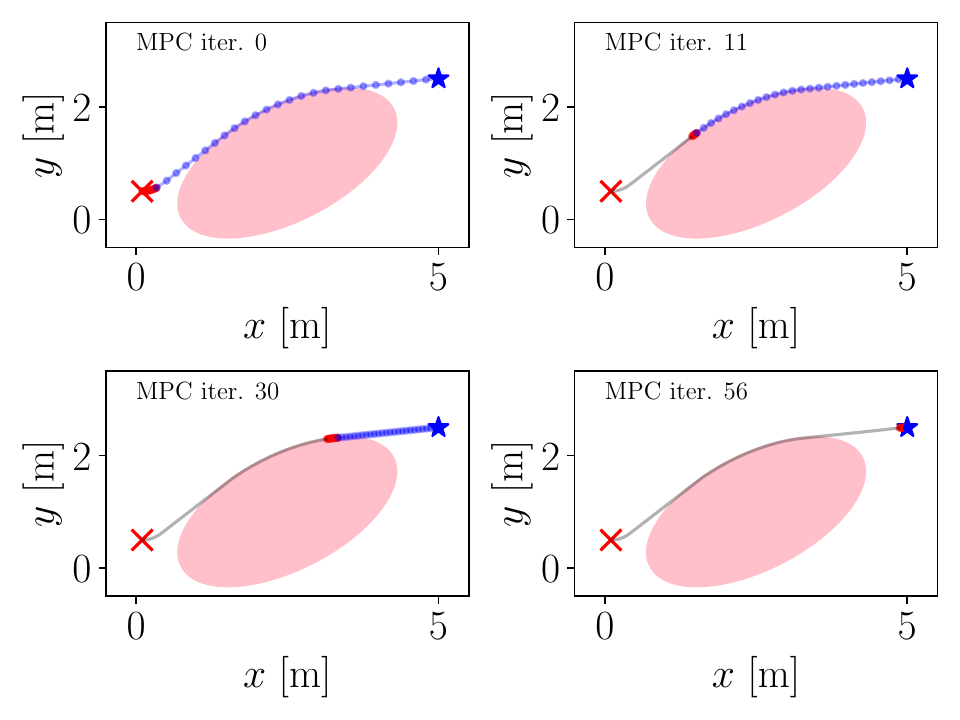}    % The printed column width is 8.4 cm.
    \caption{Four instances of the NMPC example. The red cross, the blue star, the black line, the red and the blue dot lines denote the initial and desired terminal positions, the executed trajectory, the remaining stage 1 trajectory, and the stage 2 trajectory, respectively.} 
    \label{fig:asap_2s_traj}
    \end{center}
\end{figure}

\begin{figure}[t]
    \begin{center}
    \includegraphics[width=8.4cm]{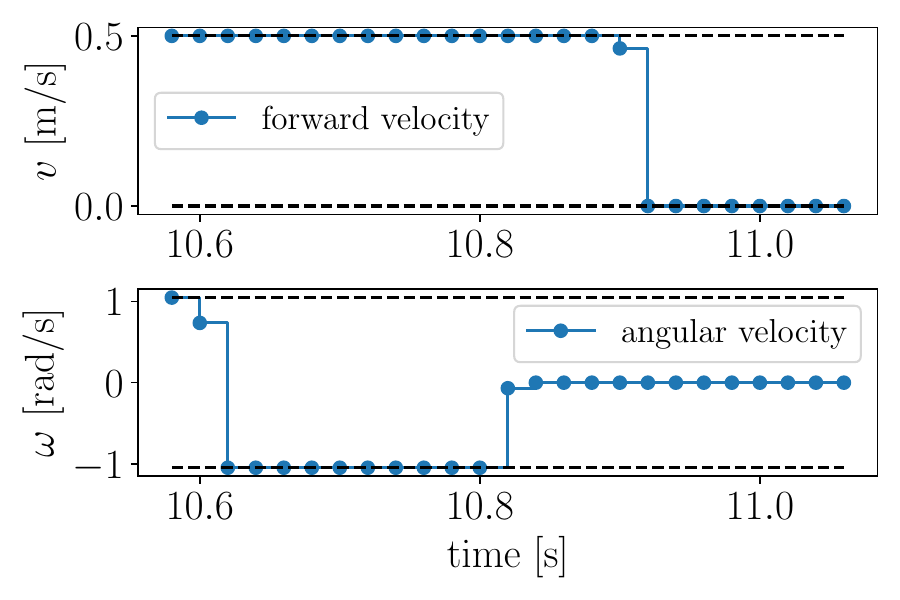}    % The printed column width is 8.4 cm.
    \caption{The control inputs of stage 1 obtained from the 56th replanning, with black dashed lines denoting the control limits.} 
    \label{fig:asap_final_u}
    \end{center}
\end{figure}

The OCP (\ref{ocp: two_stage}) is iteratively solved with current initial state $s_{\text{t0}}$ until the desired terminal state $s_{\text{tf}}$ is reached.
Fig. \ref{fig:asap_2s_traj} depicts four time instances when the current solution is available, seamlessly stitched with the executed trajectory, and concurrently initiates a new planning with the updated current state $s_{\text{t0}}$.
At the outset, the system is at $s_{\text{t0}}=[0.1\mathrm{m}, 0.5\mathrm{m}, 0\mathrm{rad}]^\top$ and initiates planning a time-optimal trajectory starting from this $s_{\text{t0}}$.
The initial planned trajectory, depicted in the top-left subplot of Fig. \ref{fig:asap_2s_traj}, encompasses a total trajectory time of 10.9191s.
Subsequently, it will continuously replan. 
As an instance, the 11th replanning takes $n_{\text{update}}=15$ control sampling times to solve, and the resulting trajectory is displayed in the top-right subplot of Fig. \ref{fig:asap_2s_traj}. 
In this scenario, the first 15 trajectory points are seamlessly stitched with the executed trajectory, which is depicted in the black line.
The remaining trajectory of stage 1 and the trajectory of stage 2 are depicted in red and blue dot lines, respectively.
The 15th on-trajectory state of stage 1 becomes the updated $s_{\text{t0}}$ for the subsequent replanning.
The total trajectory time, which is the sum of the trajectory time of the current executed trajectory and the trajectory time of the upcoming replanned trajectory, amounts to 10.9179 seconds. 
Importantly, this duration still embodies the time-optimal solution.
This process and conclusion remain consistent in the 30th (left-bottom subplot of Fig. \ref{fig:asap_2s_traj}), 56th (right-bottom subplot of Fig. \ref{fig:asap_2s_traj}), and all other replannings.
One remark is that the 56th replanning serves as an instance where the trajectory time of stage 2 is zero.
Therefore, with updated weighting factors, the two-stage approach aligns with the exponential weighting approach and stabilizes the system to the desired terminal state $s_{\text{tf}}$ relying solely on stage 1.
Fig. \ref{fig:asap_final_u} depicts the control inputs of stage 1 obtained from the 56th replanning, indicating that the system reaches the desired terminal state $s_{\text{tf}}$ at $t=10.92s$, at which point the forward velocity is zero.
This can be deemed as the time-optimal solution, as discussed in Section \ref{Sec_IV1}.

\begin{figure}[t]
    \begin{center}
    \includegraphics[width=8.4cm]{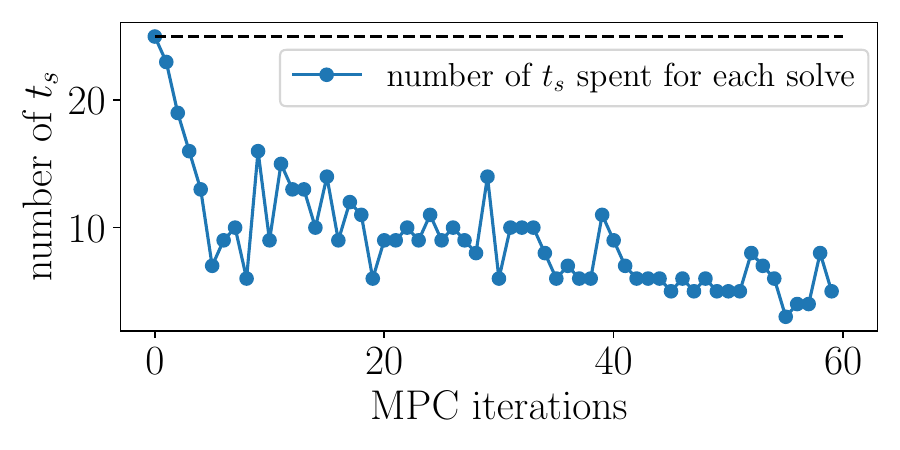}    % The printed column width is 8.4 cm.
    \caption{Number of control sampling time $t_s$ spent for each solve. The black dashed line denotes the upper bound.} 
    \label{fig:asap_2s_tcomp}
    \end{center}
\end{figure}

In total, it plans 60 times until reaching the desired terminal state $s_{\text{tf}}$.
Fig. \ref{fig:asap_2s_tcomp} illustrates the number of control sampling times spent for each solve.
In this example, each replanning takes no longer than the fixed total time of stage 1, i.e., $N_1t_s=0.5s$.
\section{Conclusion}\label{Sec_V}
We propose a two-stage approach to formulate the time-optimal point-to-point motion planning problem. 
The corresponding time-optimal OCP formulated using this two-stage approach features a fixed and low number of control steps for computational manageability.
In the first stage, it uses a fixed time grid corresponding to the control sampling time, preempting interpolation errors.
To handle fluctuating and delayed computation times in solving the time-optimal OCP, which exceed one control sampling time, we integrate the two-stage approach with the ASAP-MPC update strategy, facilitating online replanning. 
Numerical examples of autonomous navigation with collision avoidance demonstrate this integration and highlight the advantages of the two-stage approach over other alternative approaches in the literature.
The subject of further work is the implementation of the two-stage approach on embedded systems, in particular autonomous mobile robots, for real-world time-optimal motion planning tasks.
\bibliography{ifacconf} 

\end{document}